\documentclass{article}

\usepackage{PRIMEarxiv}

\usepackage[utf8]{inputenc} 
\usepackage[T1]{fontenc}    
\usepackage{hyperref}       
\usepackage{url}            
\usepackage{booktabs}       
\usepackage{amsfonts}       
\usepackage{nicefrac}       
\usepackage{microtype}      
\usepackage{lipsum}
\usepackage{fancyhdr}       
\usepackage{graphicx}       
\graphicspath{{media/}}     
\usepackage{amsmath}
\usepackage{multicol}
\usepackage{multirow}
\usepackage{graphicx}
\usepackage{float}

\title{Spatially-resolved Thermometry from Line-of-Sight Emission Spectroscopy via Machine Learning
}

\author{
   Ruiyuan Kang \\
  Department of Mechanical Engineering \\
  Khalifa University \\
  Abu Dhabi, UAE\\
  \texttt{ruiyuan.kang@ku.ac.ae} \\
   \And
  Dimitrios C. Kyritsis \\
  Department of Mechanical Engineering, RICH Center \\
  Khalifa University \\
   Abu Dhabi, UAE\\
  \texttt{dimitrios.kyritsis@ku.ac.ae} \\
     \And
  Panos Liatsis \\
  Department of Electrical Engineering and Computer Science \\
  Khalifa University \\
   Abu Dhabi, UAE\\
  \texttt{panos.liatsis@ku.ac.ae} \\
}

\begin{document}
\maketitle

\begin{abstract}
A methodology is proposed, which addresses the caveat that line-of-sight emission spectroscopy presents in that it cannot provide spatially resolved temperature measurements in nonhomogeneous temperature fields. The aim of this research is to explore the use of data-driven models in measuring temperature distributions in a spatially resolved manner using emission spectroscopy data. Two categories of data-driven methods are analyzed: (i) Feature engineering and classical machine learning algorithms, and (ii) end-to-end convolutional neural networks (CNN). In total, combinations of fifteen feature groups and fifteen classical machine learning models, and eleven CNN models are considered and their performances explored. The results indicate that the combination of feature engineering and machine learning provides better performance than the direct use of CNN. Notably, feature engineering which is comprised of physics-guided transformation, signal representation-based feature extraction and Principal Component Analysis is found to be the most effective. Moreover, it is shown that when using the extracted features, the ensemble-based, light blender learning model offers the best performance with RMSE, RE, RRMSE and R values of 64.3, 0.017, 0.025 and 0.994, respectively. The proposed method, based on feature engineering and the light blender model, is capable of measuring nonuniform temperature distributions from low-resolution spectra, even when the species concentration distribution in the gas mixtures is unknown.
\end{abstract}

\section{Introduction}
\label{section1}

Emission spectroscopy is a popular gas-thermometry technique, extensively utilized in the fields of remote sensing \cite{meerdinkPlantSpeciesSpectral2019a} and combustion diagnostics \cite{solomonFTIREmissionTransmission1988a}, among others. It naturally relies on the line-of-sight property, since a spectrum is obtained through the convolution of the emission signals along the light path. Thus, in the context of thermometry, only a single average temperature measurement, rather than a spatially resolved temperature profile, can be obtained \cite{parameswaranGasificationTemperatureMeasurement2014}. Limited by this caveat, the use of emission spectroscopy-based thermometry is constrained to scenarios involving roughly homogeneous gases.

Although advanced methods, such as tomographic spectroscopy \cite{liuInverseRadiationProblem2019} and laser-induced fluorescence \cite{eineckeMeasurementTemperatureFuel2000} are able to provide spatially resolved temperature measurements in the gaseous phase, emission spectroscopy is much simpler in terms of the hardware requirements, cheaper, and easier to use. Moreover, emission spectroscopy can be extended to hyperspectral imaging, which is a core technology in remote sensing \cite{gerhardsChallengesFuturePerspectives2019,veraverbekeHyperspectralRemoteSensing2018}. Therefore, the development of advanced estimation methods for spatial temperature distributions from line-of-sight emission spectra could support the development of a powerful thermometry tool for use in both research and development applications.

\begin{figure}[hbt!]
\centering
\includegraphics[width=.9\textwidth]{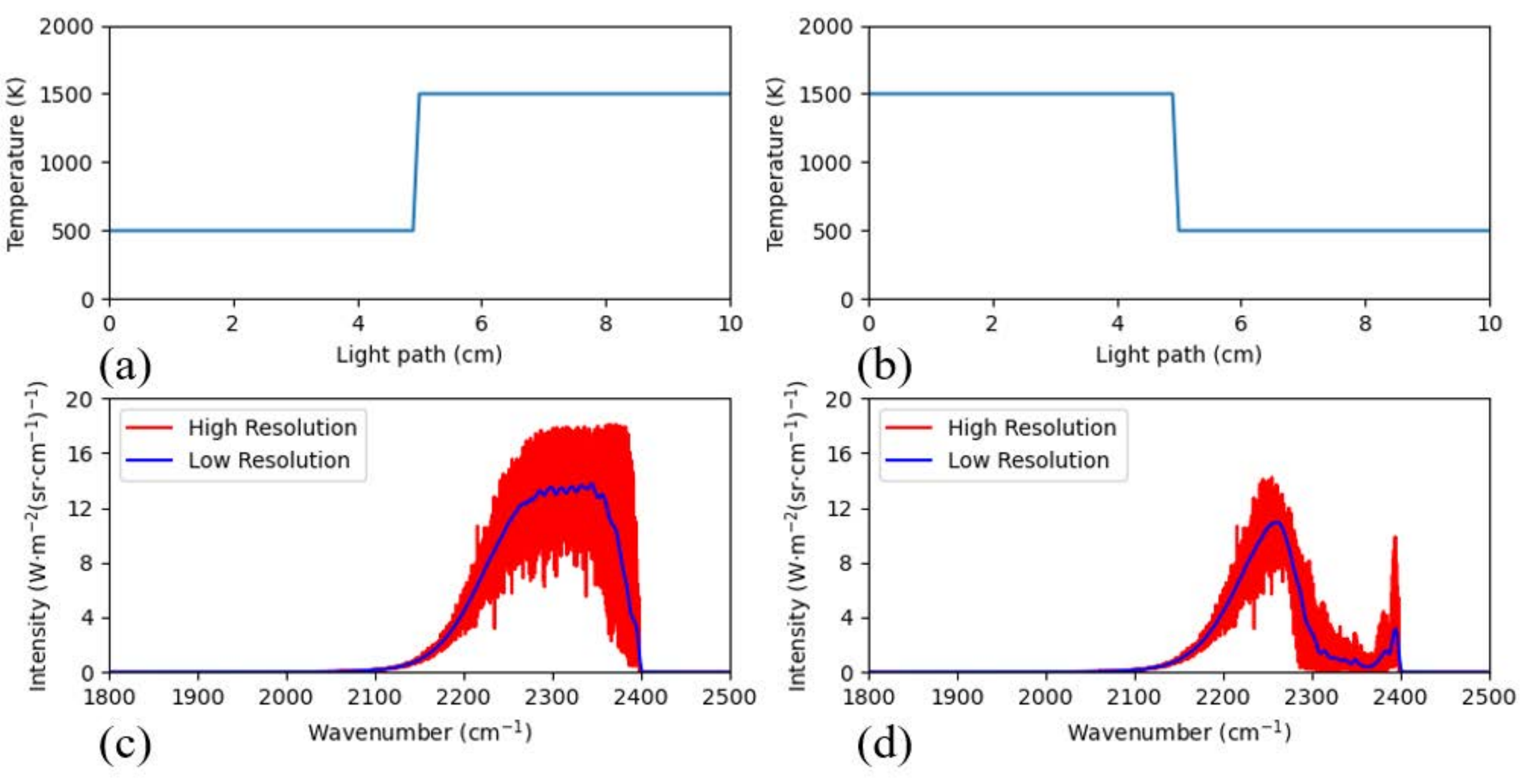}
\caption{ Comparison of spectra captured from the two ends of the same light path. (a) Temperature profile along the light path from left (500 K) to right (1500 K); (b) Temperature profile along the light path from right (500 K) to left (1500 K); (c) Emission spectrum generated from the light path in (a); (d) Emission spectrum generated from the light path in (b). Both high and low resolution spectra is visualized in (c) and (d).}
\label{fig1}
\end{figure}

The motivation behind this work is that the spatial nature of the temperature distribution is not lost during the scanning process, but rather it is encoded within the emission spectrum representation. Figure~\ref{fig1} illustrates this point. Suppose that the light path develops along a homogenous gas, where the mole fraction of CO$_2$ is 0.1, divided into two segments with temperatures of 500 K and 1500 K, respectively. Assume that two spectrometers are respectively placed at either end of the light path to capture the spectral information. As shown in Figures~\ref{fig1}(a) and (b), the temperature profiles along the light path, which will be used to capture the associated spectra, are, of course, the inverse of each other. The HITRAN Application Programming Interface (HAPI) \cite{kochanovHITRANApplicationProgramming2016} was used to simulate the spectra for each of the two cases. Figures~\ref{fig1} (c) and (d) show both the corresponding high-resolution spectra (resolution of 0.1 $cm^{-1}$), and the low-resolution version (resolution of 4 $cm^{-1}$), after the tuning of the triangular slit functions. It is evident that the two spectra are totally different in terms of appearance, irrespective of the choice of resolution. Indeed, the primary variable controlling their appearance is the ordering of two temperature segments, i.e., 500 K and 1500 K. This observation supports the hypothesis that temperature information is encoded rather than eliminated during the acquisition process of the emission spectrum. 

A pioneering idea to realize spatially resolved temperature measurements from line-of-sight data is inverse modelling by using regularization of profile smoothness \cite{kimDeterminationGasTemperature2005,songSpectralRemoteSensing2008,renTemperatureProfileInversion2014}. In this method, a radiative model is used in order to approximate the test spectrum by varying the spatial distribution of temperature. By monitoring the consistency between the estimated and test spectra, the quality of the measured spatially resolved temperature distributions can be assessed. However, since spectrum intensity does not only depend on temperature, but also the concentration of the species of interest, to avoid divergence of temperature measurement, the concentration distribution is supposed to be known in some studies \cite{kimDeterminationGasTemperature2005,renTemperatureProfileInversion2014}. Moreover, the hyperparameters of the utilized optimization algorithms often require careful adjustment, in order to allow the matching of measurements with the ground truth \cite{songSpectralRemoteSensing2008,renTemperatureProfileInversion2014}. Unfortunately, neither prior knowledge of concentration nor the ground truth of the temperature distribution, which is the task at hand in this research, are available in practice, and thus, there is no direct information to support adjustment of the associated hyperparameters. 

The inherent difficulties in extracting and interpreting the latent spatial temperature information within line-of-sight emission spectra using the physics-driven perspective, naturally suggest exploring the power of data-driven methods. A number of studies attempted the use of machine learning in the context of such measurements \cite{cieszczykDeterminationPlumeTemperature2015,renMachineLearningApplied2019}. However, they were limited in using Multilayer Perceptrons (MLP) \cite{gardnerArtificialNeuralNetworks1998}. Recent advances in machine learning (ML) and deep learning (DL), e.g., building on the paradigms of Convolutional Neural Networks (CNN) \cite{al-saffarReviewDeepConvolution2017}, and blending methods \cite{chen2012linear}, may be able to improve measurement accuracy. Moreover, the length of the acquired spectral measurements translates to a high dimensional representation of the associated information, which could be problematic, when directly used with classical ML algorithms. This problem is termed the curse of dimensionality \cite{trunkProblemDimensionalitySimple1979}, and relates to the increase in the dimensionality of the solution space, leading to sparse data coverage, which subsequently affects the quality of the ML solution. A means to tackle this is extracting lower dimensional, compact features to represent the high dimensional information, also termed as Feature Engineering (FE).

This research systematically addresses the challenge of recovering spatial temperature information from line-of-sight emission spectra measurements using data-driven modelling approaches. The contributions of the work are summarized as follows:
\begin{enumerate}
  \item A variety of feature engineering approaches were considered. Knowledge of the physical model of the process was used to transform the spectra, and next, fifteen feature groups were extracted to succinctly describe their characteristics. A wrapper-based feature selection method, i.e., utilizing an MLP as the means to quantify feature quality, was used to select the most informative features, which were found to be the combination of physics-guided transformation, signal representation-based feature extraction \cite{husainTactileSensingUsing2021}, and Principal Component Analysis \cite{abdiPrincipalComponentAnalysis2010}.
  \item Extensive exploration of classical machine learning algorithms was pursued. In addition to the MLP algorithm, a variety of methods, including Radial Basis Function Networks \cite{elanayarv.t.RadialBasisFunction1994}, Gaussian Process Regression \cite{williams2006gaussian}, etc. were attempted. In total, fifteen machine learning models were trained and tested. It was found that the best performance was obtained with the light blender model, a two-stage ensemble learning model, realized by a blending method \cite{chen2012linear}, which produced superior results, compared to the state-of-the-art \cite{cieszczykDeterminationPlumeTemperature2015,renMachineLearningApplied2019}.
  \item DL/CNN methods were explored in detail. We compared the performance of traditional ML methods coupled with feature engineering, against a number of end-to-end DL algorithms, such as Resnet \cite{heDeepResidualLearning2016} and Shuffle Net \cite{zhangShuffleNetExtremelyEfficient2017}. It was found that the combination of raw spectra inputs and CNN was not able to match the excellent performance of the combination of feature engineering and ML. This demonstrates that feature engineering can play an important role in certain application domains, including the problem under consideration in this research.
\end{enumerate}

\section{Methodology}
\label{section2}

\begin{figure}[hbt!]
\centering
\includegraphics[width=0.8\textwidth]{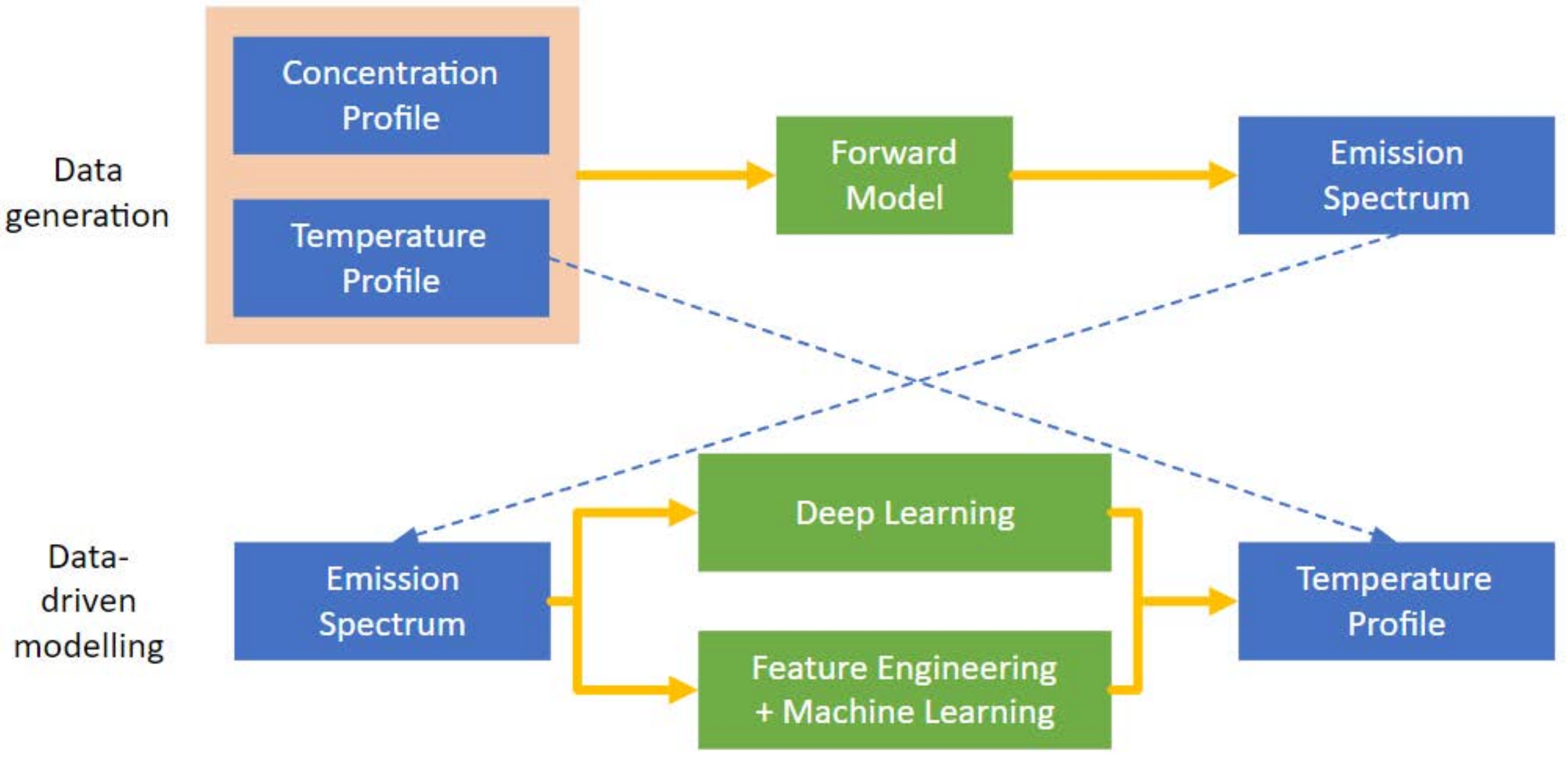}
\caption{Overview of research methodology}
\label{fig2}
\end{figure}

Figure~\ref{fig2} provides an overview of the overall system methodology. It basically includes two steps: (a) data generation, i.e., using the physical (forward) model to synthetize emission spectra, and (b) data-driven modelling, where the generated data is used to train and test the performance of FE+ML/DL models.

\subsection{Forward Model and data generation}
\subsubsection{Physical forward model}
In order to synthese spectra for the purpose of training and testing machine learning models, a forward emission spectroscopy model was constructed, based on the well-established HITEMP2010 database \cite{rothmanHITEMPHightemperatureMolecular2010}, and its simulation platform, HAPI \cite{kochanovHITRANApplicationProgramming2016}.

\begin{figure}[hbt!]
\centering
\includegraphics[width=0.5\textwidth]{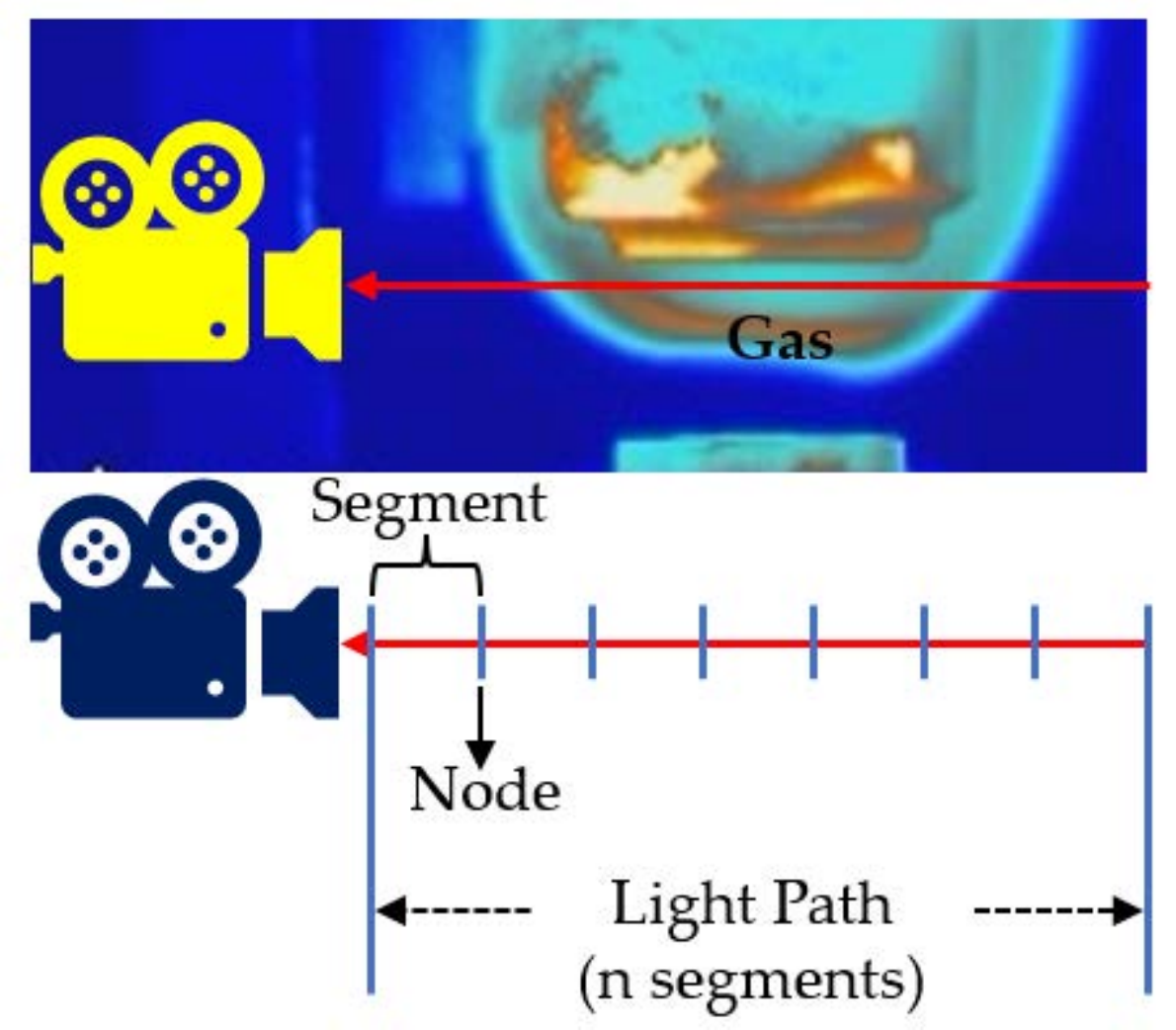}
\caption{Schematic diagram of light path division}
\label{fig3}
\end{figure}

For the purpose of generating spectra from nonuniform temperature profiles, we first discretized the light path, going through a gas cloud (such as a flame), into n segments, as shown in Figure~\ref{fig3}. Thus, nonuniform temperature and concentration profiles could be modelled within these segments.

Following the theory of spectroscopy \cite{hansonSpectroscopyOpticalDiagnostics2016a}, the intensity at the exit of a segment is the sum of the emission of the current segment and the transmission of the radiation from the previous segments:
\begin{equation}
\label{eq1}
  I_{v,o}=\epsilon_v I_{v,B}+t_vI_{v,i}
\end{equation}
where $I$, $\alpha$, $t$ are the intensity, emissivity, and transmissivity, respectively. The subscripts $v$,$B$,$i$, and $o$ relate to the frequency, Black-body radiance, incident and outlet nodes of the segment, respectively. According to the Beer-Lambert law \cite{hansonSpectroscopyOpticalDiagnostics2016a}, transmissivity can be expressed by:
\begin{equation}
\label{eq2}
  t_v=exp(-k_vl)
\end{equation}
where $k_v$ is the absorption coefficient $(cm^{-1})$, and $l$ is the length of the light path segment (cm). The sum of transmissivity $t$ and absorptivity $\alpha$ is equal to 1, when scattering and reflection are neglectable, i.e.,
\begin{equation}
\label{eq3}
  t_v+\alpha_v=1
\end{equation}

Let us consider that the gas cloud is at thermal equilibrium. By following Kirchhoff's law of thermal radiation \cite{hansonSpectroscopyOpticalDiagnostics2016a} that emissivity equals to absorptivity, and substituting Eqs.\ref{eq2}-\ref{eq3} into Eq.\ref{eq1}, we get: 
\begin{equation}
\label{eq4}
  I_{v,o}=(1-exp(-k_vl))I_{v,B}+exp(-k_vl)I_{v,i}
\end{equation}

Moreover, the black-body radiance can be modeled by Plank’s law \cite{bergmanFundamentalsHeatMass2011}:
\begin{equation}
\label{eq5}
  I_{v,B}=\frac{2hv^3}{c^2(exp(\frac{hv}{k_BT})-1)}
\end{equation}
where h is the Plank constant, $k_B$ is the Boltzmann constant, $c$ is the light speed, and $T$ is the temperature. Then, the absorption coefficient, $k_v$, of gas species $j$ is given by:
\begin{equation}
\label{eq6}
  k_{v,j}=s_{v,j}(T)\phi_v(T,p)\frac{pX_j}{k_BT}
\end{equation}

where s is the line intensity per molecule ($(cm)^{-1}/(molecule*cm^{-2})$), which is a function of temperature, $\phi_v$ is the Voigt profile \cite{rostonExactAnalyticalFormula2005}, which is a function of both pressure and temperature, and $p$ is the local pressure ($Pa$), and $X$ is the mole fraction. The total absorption coefficient can be simplified as the sum of the absorption coefficients of each species:
\begin{equation}
\label{eq7}
  k_v=\sum k_{v,j}
\end{equation}

Although high resolution spectra are better in providing precise information, low-resolution spectra, which are affected by blurring due to line broadening and the slit function of the instrument, are the norm in engineering measurements. A triangular function was used to simulate the instrument slit function as follows:
\begin{equation}
\label{eq8}
B(x) =
\begin{cases}
\begin{aligned}
\frac{1-\frac{abs(v-v_c)}{w}}{w}\qquad & abs(v-v_c) \le w \\
0\qquad & abs(v-v_c)>w
\end{aligned}
\end{cases}
\end{equation}
where $w$ is the wing of the triangular function, which is set to 10 $cm^{-1}$, and $v_c$ is the center of the triangular function, in this work, the resolution of spectrum is 4 $cm^{-1}$.

By considering the aforementioned equations, a number of factors can be identified that hinder retrieval of spatially resolved temperature measurements. First, as observed in Eq.\ref{eq6}, both temperature and species concentration affect the appearance of the spectrum. Thus, lack of prior information in regards to the species concentration increases the difficulty in retrieving the varying temperature profile along the light path. Second, as shown in Eq. \ref{eq7}, in the case of gas species mixtures, the total absorption coefficient is affected by the spectral characteristics of all species, thus, posing a further complication in estimating the temperature measurement information. Third, blurring due to various factors, including the instrument slit function of Eq. \ref{eq8}, reduces the quality of spectral information, increasing the ambiguity in temperature measurement. Last, temperature profile non-uniformity and lack of prior information in respect to the range of the temperature distribution complicate the extraction of patterns, governing the spectra.

\subsubsection{Data generation}
In this research, we focus on the paradigm of combustion. This is a suiTablescenario for spatially resolved temperature measurements with emission spectroscopy, because the high temperatures of the combustion process provide sufficient emission signals. Moreover, most practical combustion phenomena naturally possess nonuniform temperature distributions, which, in turn, require spatially resolved temperature measurements. In the data generation, the conditions, which occur in combustion were replicated, however, more complex situations were also considered. This was achieved by varying each of the four parameters, which affect the appearance and quality of the spectra, subsequently affecting temperature profile estimation, i.e., uncertainties in concentration, mixture of species, nonuniformities and range of temperature distribution, and spectral resolution. This methodology was followed to systematically assess the capabilities of ML/DL algorithms in tackling the temperature measurement problem. The detailed data configuration is described in the following paragraphs.

\begin{figure}[hbt!]
\centering
\includegraphics[width=0.9\textwidth]{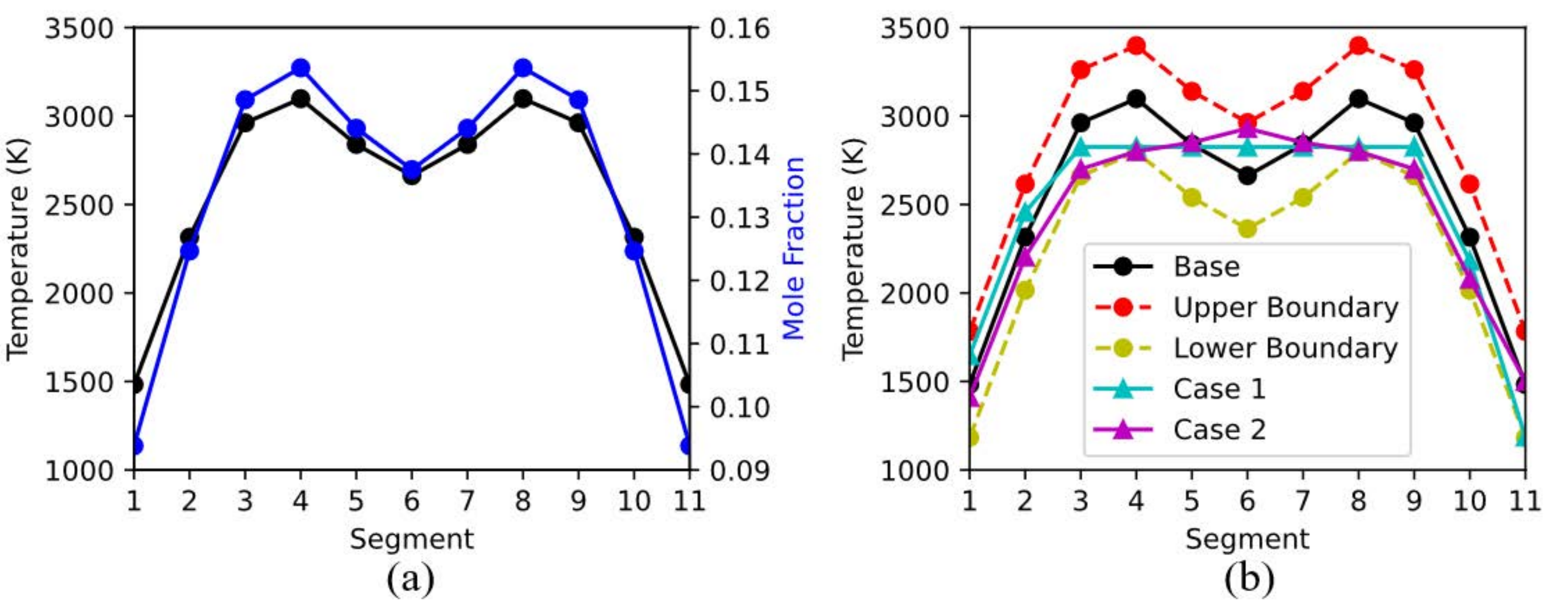}
\caption{Temperature and concentration profiles. (a) Baseline-- ideal profile, (b) Examples of temperature profiles, generated through random variation of the associated parameters.}
\label{fig4}
\end{figure}

The gases considered were CO, CO$_2$ and H$_2$O, because of their importance in combustion applications. As previously mentioned, such gas mixtures also complicate temperature measurement estimation. Following \cite{renMachineLearningApplied2019}, the wavebands selected were in the range of 1800-2500 $cm^{-1}$, which covers the emission bands of these three substances. The light path was set to be ten cm and divided into eleven segments, i.e., there were twelve nodes, with each segment being less than one cm in length. The relatively small segment length and the number of the segments suffice to validate the feasibility of measuring spatially resolved temperature measurements. Indeed, the number of segments may vary according to the requirements of the application. The dual-peak Gaussian function, commonly encountered in flames \cite{goldensteinScannedwavelengthmodulationSpectroscopyMm2014}, was used as the base profile herein. The ideal dual-peak Gaussian function is given by:
\begin{equation}
\label{eq9}
\rho_{DG,j}=exp(-n(\frac{j-n}{4\sigma})^2)+exp(-n(\frac{3(j-n)}{4\sigma})^2)
\end{equation}
where $\rho_{DG,j}$ is the density of dual-peak Gaussian profile at segment $j$, $n$ is the number of segments, $\sigma$ is the spread, which was empirically set to 16. The values of $\rho_{DG,j}$ were normalized in the range of (0,1). The feasible ranges of temperature and mole fraction can be obtained from the normalized density as follows: 
\begin{equation}
\label{eq10}
T_{DG,j}=\rho_{DG,j,norm}(T_{max}-T_{min})+T_{min}
\end{equation}
\begin{equation}
\label{eq11}
X_{DG,j}=\rho_{DG,j,norm}(X_{max}-X_{min})+X_{min}
\end{equation}

where $\rho_{DG,j,norm}$ is the normalized basis height, $T_{DG,j}$ and $X_{DG,j}$ are, respectively, the temperature and mole fraction at segment j, $T_{max}$, $T_{min}$ and $X_{max}$, $X_{min}$ are, respectively, the maximum and minimum of temperature range and mole fraction range. The corresponding temperature range was set to 1500 K-3100 K, i.e., covering a broad interval of 1600 K, while the mole fraction range was set to 0.095-0.15, with examples of ideal generated profiles demonstrated in Figure~\ref{fig4} (a).

Next, the values of temperature and gas concentration in each segment were randomly varied in the ranges of ±300 K and ±0.015, respectively. Examples of temperature profiles generated by introducing such variations are shown in Figure~\ref{fig4} (b). It is evident that such variations have a significant impact on the shape of generated profiles. For instance, see Figure~\ref{fig4} (b), cases 1 and 2, which are generated from such profile configurations , do not exhibit the traditional shape of the dual peak Gaussian function. Rather, they resemble the shapes of trapezoidal and parabolic profiles, respectively. It should be emphasized that such profiles are also commonly observed in boundary-layer flow and flames \cite{liuMeasurementNonuniformTemperature2007,changSupersonicMassfluxMeasurements2011}. Given the light path discretization, a large variety of profile patterns can be generated, leading to higher spectral pattern complexity, thus increasing the difficulty in obtaining spatially resolved temperature measurements. In our experiments, we considered a spectral resolution of 4 $cm^{-1}$, although a resolution of 0.1 $cm^{-1}$ is practically realizable. The purpose behind this choice is that a lower resolution leads to higher spectral generation speed, reduced instrument costs, and also it increases the challenge of obtaining temperature measurements.

\begin{figure}[hbt!]
\centering
\includegraphics[width=0.5\textwidth]{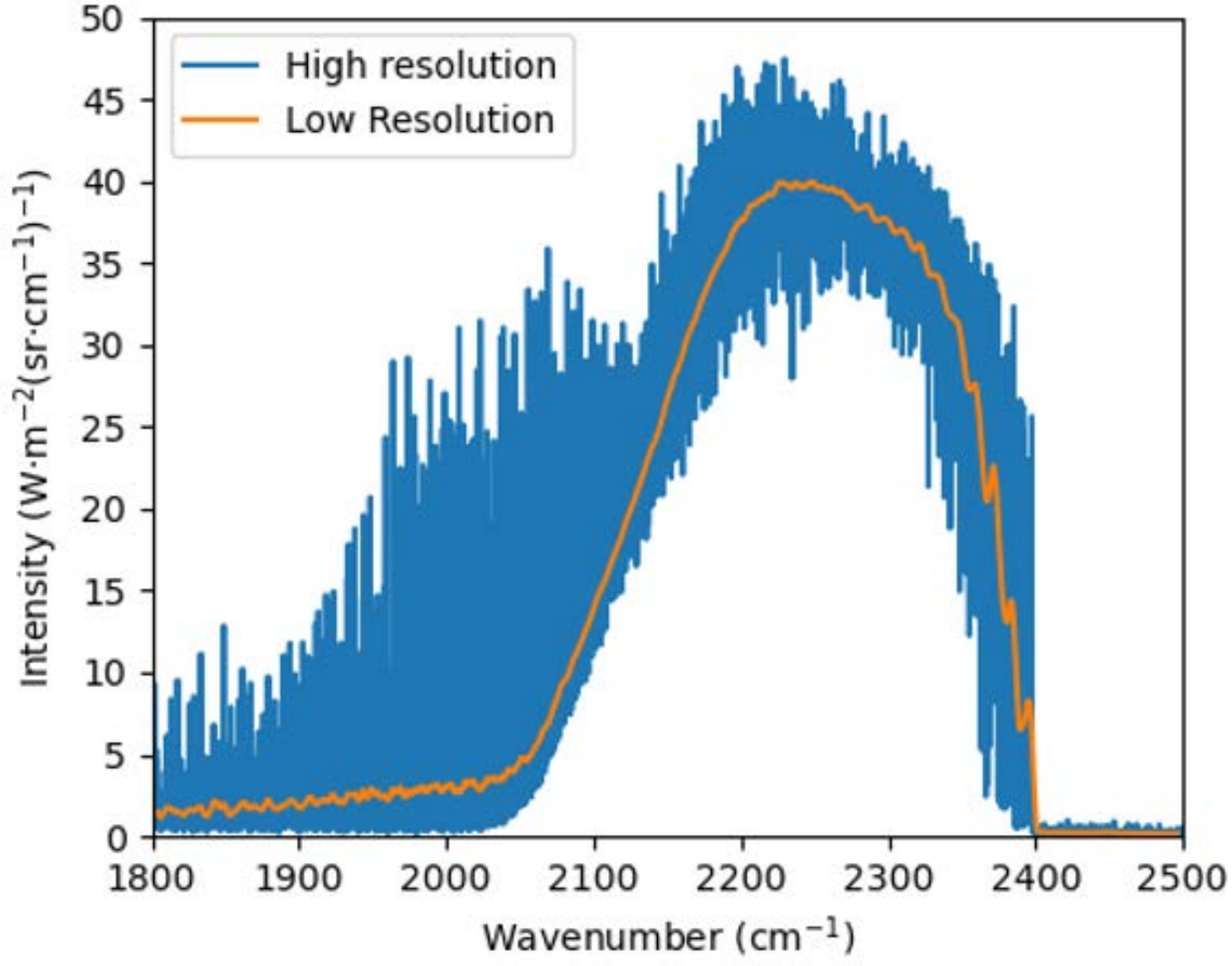}
\caption{A sample of low-resolution spectrum.}
\label{fig5}
\end{figure}

In total, 28,000 spectra were generated. A sample profile is shown in Figure~\ref{fig5}. Such a low-resolution spectrum comprises of 6799 emission lines. Compared to the high-resolution spectra of 0.1 $cm^{-1}$, also shown in the Figure, it is clear that low-resolution spectra is affected by blurring and is not able to capture finer spectral details.

\subsection{Data-driven modelling}
In the context of solving the inverse problem of line-of-sight emission spectroscopy, temperature parameters along the light path were estimated from the generated spectral dataset via data-driven modeling. Two types of data-driven modeling methods were considered: (i) feature engineering and classical machine learning algorithms; and (ii) raw spectra and CNN. 

\subsubsection{Feature engineering}
Despite the substantial size of the generated spectra samples, the high dimensionality of the spectra, i.e., 6799 emission lines, would translate to a sparse sampling of the problem space, thus posing a challenge for machine learning algorithms. Therefore, feature engineering was performed on the raw spectra inputs, so as to extract informative features, thus reducing redundancy and problem dimensionality. First, the physics-guided, logarithmic transformation was used, so as to reduce the degree of nonlinearity of the mapping between spectra and temperature. Next, two types of feature extraction methods were used, i.e., extracting statistical and signal representation features, respectively. Principal Component Analysis (PCA) \cite{abdiPrincipalComponentAnalysis2010} was subsequently used to remove noise, and further reduce the dimensionality of features. In the following subsections, we briefly introduce the feature engineering operations.

\textbf{Logarithmicic transformation.} According to Eq.\ref{eq4} in the physical forward model, the absorption coefficient which contains the information of temperature is affected by an exponential function. This introduces nonlinearities in the associated temperature-spectrum mapping, which need to be resolved through the machine learning modelling. Applying the logarithmicic transformation \cite{keeneLogTransformationSpecial1995} to the raw signals, assists by decreasing the mapping nonlinearities. This process is termed physics-guided transformation, since it is inspired by radiation physics. 

\textbf{Statistical features.} Statistical features, such as first-, second- and higher-order statistics are often used to describe high-dimensional data \cite{golgiyazArtificialNeuralNetwork2019,ogrenDevelopmentVisionbasedSoft2018}. The list of time domain statistical parameters estimated from the emission spectra is shown in Table~\ref{table1}. Moreover, the Fast Fourier transform (FFT) was applied on the spectra, and extracted frequency domain features, as shown in Table~\ref{table2}. In total, 38 features were obtained, 23 from the time domain, with the remaining ten obtained from the frequency domain. In addition to extracting these features by considering the entire waveband (1800-2500 $cm^{-1}$), it is also possible to consider these features from the characteristic bands of the chemical substances. By characteristic band, we refer to the part of the spectral range, where the signal of one mixture component is dominant, with negligible contributions originating from the other components. For H$_2$O, CO and CO$_2$, such characteristic bands are the ranges of (1800 $cm^{-1}$, 1890 $cm^{-1}$), (2100 $cm^{-1}$, 2190 $cm^{-1}$), and (2310 $cm^{-1}$, 2400 $cm^{-1}$), respectively. Thus, three further set of 38 statistical features were also extracted from each of the three characteristic bands.

\begin{table}[]
\centering
\caption{Time domain features of the emission spectra}
\label{table1}
\begin{tabular}{|l|l|}
\hline
\textbf{Features} & \textbf{Defination} \\ \hline
Mean & $\bar{I}=\frac{1}{n}\sum_{i=1}^{n}I_i$, \\
 & $i$ s the index of spectral signal, $n$=6799 \\ \hline
Maximum & $I_{max}=max(I_i)$ \\ \hline
Minimum & $I_{min}=min(I_i)$ \\ \hline
Quartile  1 & $I_{0.25*n}$, $I$ in ascending order\\ \hline
Quartile2  (median) & $I_{0.5*n}$, $I$ in ascending order \\ \hline
Quartile  3 & $I_{0.75*n}$, $I$ in ascending order \\ \hline
Interquartile  range & $I_{0.75*n}-I_{0.25*n}$ \\ \hline
Standard  deviation & $S=\sqrt{\frac{1}{n}\sum_{i=1}^{n}(I_i-\Bar{I})^2}$ \\ \hline
Variance & $S^2$ \\ \hline
Skewness & $\frac{1}{n}\sum_{i=1}^{n}(\frac{I_i-\Bar{I}}{S})^3$ \\ \hline
Kurtosis & $\frac{1}{n}\sum_{i=1}^{n}(\frac{I_i-\Bar{I}}{S})^4$ \\ \hline
Inverse  coefficient of variation & $\Bar{I}/S$ \\ \hline
Peak to  peak & $I_{max}-I_{min}$ \\ \hline
Zero cross rate & Number of signal sign changes \\ \hline
Root  mean square & $RMS_I=\sqrt{\frac{1}{n}\sum_{i=1}^{n}I_i^2}$ \\ \hline
Crest  factor & $I_{max}/RMS_I$ \\ \hline
Root mean square of difference & $S=\sqrt{\frac{1}{n-1}\sum_{i=1}^{n-1}(I_{i+1}-I_i)^2}$ \\ \hline
Root mean square of difference reciprocal & $S=\sqrt{\frac{1}{n-1}\sum_{i=1}^{n-1}\frac{1}{(I_{i+1}-I_i)^2}}$ \\ \hline
Mean of  magnitude & $\frac{1}{n}\sum_{i=1}^{n}|I_i|$ \\ \hline
Difference  variance & $\frac{1}{n-1}\sum_{i=1}^{n-1}|I_{i+1}-I_i|$ \\ \hline
Sum of  difference & $\sum_{i=1}^{n}(I_{i+1}-I_i)$ \\ \hline
Shannon entropy of spectrum & $-\sum_{i=1}^{n}I_i^2\log{I_i^2}; 0\log0=0$ \\ \hline
Log energy entropy of spectrum & $\sum_{i=1}^{n}\log{I_i^2}; \log0=0$ \\ \hline
\end{tabular}%
\end{table}

\begin{table}[]
\centering
\caption{ Frequency domain features of the emission spectra}
\label{table2}
\begin{tabular}{|l|l|}
\hline
\textbf{Features} & \textbf{Defination} \\ \hline
Average of FFT intensity & $\bar{I}=\frac{1}{n}\sum_{i=1}^{n}I_{F,i}$, \\
 & $I_F$ is the signal in the FT domain,\\
 & $i$ is the signal index in FT domain, $n$=6799 \\ \hline
Average of FFT magnitude & $\frac{1}{n}\sum_{i=1}^{n}|I_{F,i}|$ \\ \hline
Average of FFT power & $\frac{1}{n}\sum_{i=1}^{n}I_{F,i}^2$ \\ \hline
Maximal power & $I_{max}=max(I_{F,i}^2)$ \\ \hline
Minimal power & $I_{min}=min(I_{F,i}^2)$ \\ \hline
Shannon entropy of FFT & $-\sum_{i=1}^{n}I_{F,i}^2\log{I_{F,i}^2}; 0\log0=0$ \\ \hline
Log energy entropy of FFT & $\sum_{i=1}^{n}\log{I_{F,i}^2}; \log0=0$ \\ \hline
Maximal magnitudes & The first 6 maximum magnitudes of the FFT \\ \hline
Minimal magnitude & $min(|I_{F,i}|)$ \\ \hline
Average phase & $\bar{I}=\frac{1}{n}\sum_{i=1}^{n}A_{F,i}$ \\ \hline
\end{tabular}%

\end{table}

\textbf{Signal representation features.} In \cite{husainTactileSensingUsing2021}, the coefficients of polynomial basis functions were used to represent intensities in patches of a tomographic image. A similar idea was used in this research, where the spectra was decomposed into windows, each approximated by the coefficients of a basis function. These coefficients are termed signal representation features, and provide a direct description of the local shape of the spectrum, rather than the abstract representation provided by the statistical features.

Two window lengths ($n$) were explored, i.e., 20 and 50 emission lines. The reason behind the choice of relatively small window lengths is based on the assumption that signal representation approaches are able to better approximate the spectral signal pattern inside a small window. However, if too small a window length is used, there are no substantial savings in terms of spectrum representation, compared to the raw signals. A variety of basis functions were used to approximate the spectra, i.e., polynomial, sinusoidal, exponential and power. We chose to experiment with polynomials of up to 3$^{rd}$ order, while only first-order basis functions were considered for the remaining approximations. The benefit of doing so is that the raw spectral information within each window can be represented by up to four coefficients, thus resulting in compact signal representation, thus tackling the curse of dimensionality.

The polynomial basis function approximation is given by:
\begin{equation}
\label{eq12}
  \hat{I}_{x,P}=\sum_{k=0}^{n}a_kx^k
\end{equation}
where $\hat{I}_{x,P}$ is the fitted spectral intensity at index $x$, obtained through polynomial fitting, $a_k$ is the coefficient for the kth power of $x$, $x$ is the local sample index inside the window, i.e., it varies from 1 to 20, or 50, depending on the window size, $n$ is the highest order of the polynomial approximation. In these experiments, $n$=1, 2, 3. 

Next, the first-order sinuisoidal approximation, described by the four coefficients, $c,q_1,q_2$ and $f$, is given by: 
\begin{equation}
\label{eq13}
  \hat{I}_{x,F}=c+q_1cos(fx)+q_2sin(fx)
\end{equation}

where $\hat{I}_{x,F}$ is the fitted spectral intensity at index $x$, obtained through sinusoidal fitting, $c$ is the constant term, $q_1$ and $q_2$ are the magnitudes of the cosine and sine terms, respectively, and $f$ is the frequency of the cosine and sine terms.

Similarly, the first-order exponential approximation, described by the four coefficients, $m_1,m_2,b_1,b_2$, is given by: 
\begin{equation}
\label{eq14}
  \hat{I}_{x,E}=c+m_1exp(b_1x)+m_2exp(b_2x)
\end{equation}
where $\hat{I}_{x,E}$ is the fitted spectral intensity at index $x$, obtained through exponential fitting. 

Last, the first-order power approximation, described by the three coefficients, $z,j$ and $c$, is given by:
\begin{equation}
\label{eq15}
  \hat{I}_{x,PW}=zx^j+c
\end{equation}
where $\hat{I}_{x,PW}$ is the fitted spectral intensity at index $x$, obtained through power fitting, $z$ is the magnitude of the power term, $j$ is the power coefficient, and $c$ is the constant term.

\textbf{Principal Component Analysis (PCA).} PCA is a broadly used technology for feature selection and dimensionality reduction. It rebuilds the original feature space by a group of orthogonal bases which correspond to the directions of most variance of the features, and the original features are thus decomposed and represented in the bases’ direction with the decreasing order of magnitude of variance. Consequently, the order of the new features demonstrates the importance and informativity of the feature, and the features with very low orders can be regarded as noise or unimportant information.
\subsubsection{Machine Learning/Deep Learning}
A variety of traditional machine learning models were considered in the experiments, including Multi-Layer Perceptron (MLP) \cite{gardnerArtificialNeuralNetworks1998}, Gaussian Process Regression (GPR) \cite{williams2006gaussian}, Support Vector Regression (SVR) \cite{nobleWhatSupportVector2006}, Radial Basis Function Network (RBFN) \cite{elanayarv.t.RadialBasisFunction1994}, and tree-based ensemble learning algorithms \cite{sagiEnsembleLearningSurvey2018}, i.e., Random Forests \cite{borupTargetingPredictorsRandom2022} and Boosted Trees \cite{chen2016xgboost}. In addition to utilizing these methods to estimate the spatially-resolved temperature profiles, the ensemble learning framework of the blending method \cite{chen2012linear} was applied to fuse together the individual model estimates.

\begin{figure}[hbt!]
\centering
\includegraphics[width=0.7\textwidth]{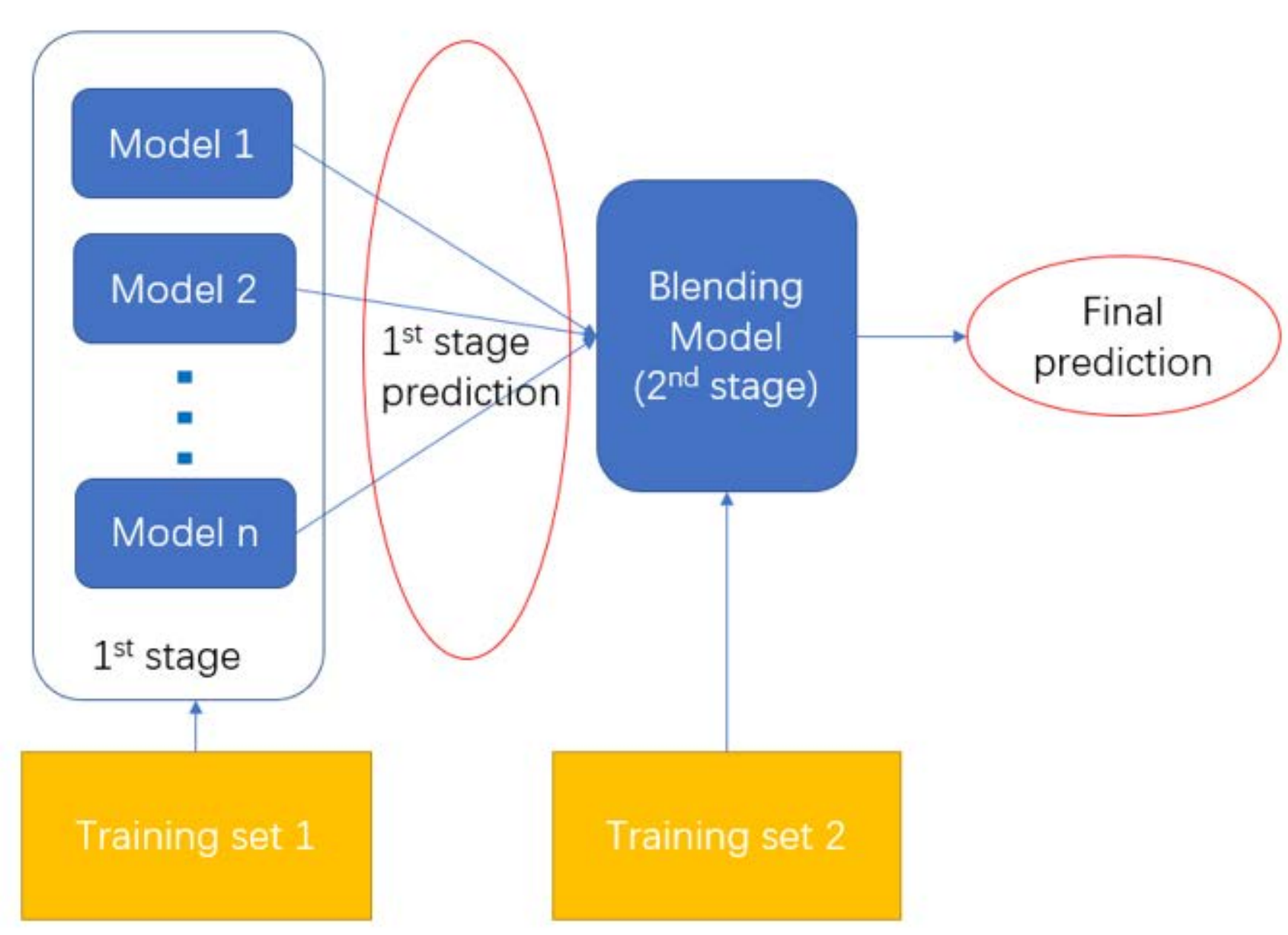}
\caption{Schematic of the blending method.}
\label{fig6}
\end{figure}

The blending method consists of two stages, as shown in Figure~\ref{fig6}. The first stage models are the weak learners, i.e., the machine learning models used to tackle the problem individually. The second stage of the process is the blending model or meta learner, which provides the final estimates, using the primary estimates from the weak learners. The meta learner can be a linear (e.g., weighted average) or a nonlinear model (e.g., MLP). It should be noted that when training the meta learner, the training set needs to be different from that of the weak learners. Otherwise, data leakage \cite{samalaHazardsDataLeakage2020} could occur, i.e., the meta learner could be biased towards the weak learner, which performs the best in the original training set. To ensure that training was properly done, a validation set was used to train the meta learners.

In order to have a means of comparison for the combination approach of feature engineering and machine learning algorithms, the raw spectra was processed using state-of-the-art CNN architectures \cite{al-saffarReviewDeepConvolution2017}. CNN is well-known for its ability to automatically extract features, thus alleviating the need for feature engineering. The CNN architectures investigated in this research are VGG series \cite{simonyanVeryDeepConvolutional2015}, Resnet series  \cite{heDeepResidualLearning2016}, Inception series \cite{cholletXceptionDeepLearning2017,szegedyGoingDeeperConvolutions2014}, Squeeze net \cite{iandolaSqueezeNetAlexNetlevelAccuracy2016} and Shuffle net \cite{zhangShuffleNetExtremelyEfficient2017}.  

As these CNNs were originally designed for image classification tasks, some modifications were made prior to using them for the regression task of measuring spatially resolved temperatures. The main changes were as follows: (1) batch normalization and dropout were removed from the networks, as batch normalization shifts the center and standard deviation of the normalization operation with respect to the varied sampling of data, thus leads to high bias and variance for the regression task, (2) the last 399 spectral intensities were discarded, and the remaining 6400 intensities were reshaped to a square image of 80*80. The architectures utilized and modified are from the open-sourced repository: Pytorch-Cifar100 \url{https://github.com/weiaicunzai/pytorch-cifar100}.

\section{Results and Discussion}
\subsection{Application of feature engineering and machine learning}
\subsubsection{Feature engineering}
First, we report the application of feature engineering and machine learning methods. The sequence of processing for feature engineering was as follows: (i) physics-guided transformation, (ii) statistical/representation-based feature extraction, (iii) PCA.

After the physics-guided (logarithmicic) transformation, thirty-eight statistical features were extracted from the entire spectrum, which were used as the first group of features. Moreover, for each of the three characteristic bands of H$_2$O, CO and CO$_2$, a further thirty-eight features were extracted, respectively, which were then used as the second group of features. The hypothesis, which we wished to explore is whether the information contained within these two sets of features is distinct and complementary. Thus to obtain a comprehensive description of the information of the entire spectrum, the two feature groups were merged together, resulting to a third feature group. 

In the case of representation-based features, each spectrum signal was divided into windows of either 20 or 50 spectral samples, followed by their approximation using the four sets of basis functions. The approximation coefficients were then used to represent the information of the spectrum. By doing so, twelve groups of signal representation features were obtained, as shown in Table~\ref{table3}. The notation used for the reporting of the signal representation features involves their order and window length, e.g., 3$^{rd}$-20 means third order approximation was used to fit the window data, containing 20 samples.

\begin{table}[]
\centering
\caption{Comparison of fitting quality of Signal representation features}
\label{table3}
\begin{tabular}{|cc|c|c|c|}
\hline
\multicolumn{2}{|c|}{\textbf{Fitting basis function}} & \textbf{Feature number} & \textbf{MSE} & \textbf{R} \\ \hline
\multicolumn{1}{|c|}{\multirow{6}{*}{Polynomial fitting}} & 3$^{rd}$-20 & 1356 & \textbf{1.00E-05} & \textbf{1} \\ \cline{2-5} 
\multicolumn{1}{|c|}{} & 3$^{rd}$-50 & 540 & 9.20E-05 & 0.9999 \\ \cline{2-5} 
\multicolumn{1}{|c|}{} & 2$^{nd}$-20 & 1107 & 3.80E-05 & \textbf{1} \\ \cline{2-5} 
\multicolumn{1}{|c|}{} & 2$^{nd}$-50 & 405 & 0.0003 & 0.9997 \\ \cline{2-5} 
\multicolumn{1}{|c|}{} & 1$^{st}$-20 & 678 & 0.0001 & 0.9999 \\ \cline{2-5} 
\multicolumn{1}{|c|}{} & 1$^{st}$ -50 & 270 & 0.0011 & 0.999 \\ \hline
\multicolumn{1}{|c|}{\multirow{2}{*}{Sinusoidal fitting}} & 20 & 1356 & 0.3745 & 0.7623 \\ \cline{2-5} 
\multicolumn{1}{|c|}{} & 50 & 540 & 0.0002 & 0.9997 \\ \hline
\multicolumn{1}{|c|}{\multirow{2}{*}{Exponential fitting}} & 20 & 1356 & - & - \\ \cline{2-5} 
\multicolumn{1}{|c|}{} & 50 & 540 & - & - \\ \hline
\multicolumn{1}{|c|}{\multirow{2}{*}{Power fitting}} & 20 & 1017 & 9.50E-05 & 0.9999 \\ \cline{2-5} 
\multicolumn{1}{|c|}{} & 50 & 405 & 0.0005 & 0.9995 \\ \hline
\end{tabular}
\end{table}

\begin{figure}[hbt!]
\centering
\includegraphics[width=0.8\textwidth]{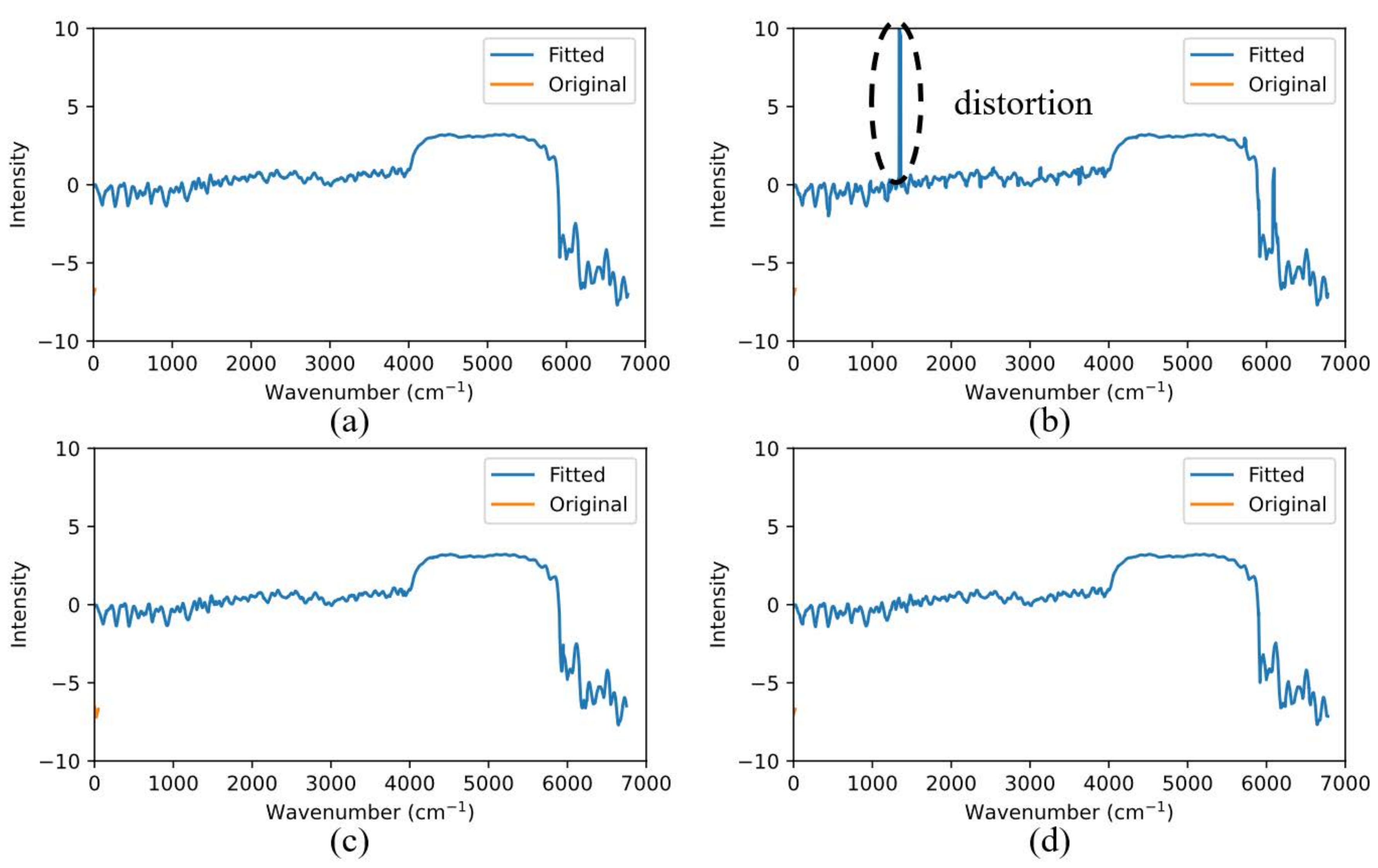}
\caption{An example of spectrum approximation using: (a) 3$^{rd}$ order polynomials and a window length of 20; (b) 1$^{st}$ order sinusoidal basis functions and a window length of 20; (c) 1$^{st}$ order sinusoidal basis functions and a window length of 50; (d) 1$^{st}$ order power basis functions and a window length of 20.}
\label{fig7}
\end{figure}

In Table~\ref{table3}, we compared the quality of approximation of the various basis functions by using the Mean Square Error (MSE) and Pearson’s correlation coefficient between the original physics-transformed spectra and their reconstructions using signal representation features. In general, the majority of methods offer high quality approximation, as demonstrated in the Tableand Figure~\ref{fig7}, except for the exponential, and sinusoidal basis with a window length of 20 samples. This is because the exponential basis functions are not able to reconstruct the spectrum as some of the coefficients become excessively large, leading to overflowing of the exponential function. In the case of sinusoidal basis with a window length of 20, the values of $q$ become very large in the case of smaller windows, thus leading to distortion. This is evident when comparing Figure~\ref{fig7}(b), where a spike is observed, in the case of a window with length of 20 samples, against Figure~\ref{fig7}(c)), when a window of 50 samples is used. Apart from these two exceptions, the observation is that the higher the order of the fit, and the smaller the window length, the higher the quality of approximation. Moreover, the polynomial basis function approximation is the most accurate among all basis function, for the same number of approximation coefficients. For instance, 2$^{nd}$ order polynomials vs 1$^{st}$ order power, and 3$^{rd}$ order polynomials vs 1$^{st}$ order sinusoidal approximation. When comparing between polynomial approximations, the case of 3$^{rd}$ -50 has comparable accuracy to the case 2$^{nd}$ -20, while the former only has about half the number of features of the latter, and is much better than the case of 1$^{st}$-20, although the latter has a larger number of features.

Moreover, PCA was applied on all feature groups. Multiple variant feature groups were derived from the original feature group by gradually eliminating the least informative PCA-transformed features. For example, using third-order polynomials and a window size of 50 resulted to 540 features, as the 6799 spectral inputs were divided into 135 windows, each window represented by four features. Following the application of PCA, all 540 PCA-transformed features were considered as a variant feature group, and subsequently reduced in order of decreasing variance to multiples of one hundred, i.e., 500,400,300, etc. 

Finally, a wrapper-based feature selection method \cite{elaboudiReviewWrapperFeature2016} was used to select the optimal feature groups, where the MLP was used as the referee, and the performance of the feature groups was assessed by its ability in approximating the spatial temperature profiles. The temperature values were normalized in the range of (0,1) using the boundary values in each segment. In principle, this operation can improve the quality of learning. The spectra dataset was divided into training, validation, and test datasets with ratios of 70\%, 15\% and 15\%, respectively. The optimal performances achieved by the various feature groups are summarized in Table~\ref{table4}. The Tableprovides information regarding the size of the original feature vector, the optimal feature vector size following the application of PCA, the optimal number of neurons in the hidden layer of the MLP, and the Mean Square Error (MSE) of training and validation sets on the normalized temperature profiles.

\begin{table}[]
\centering
\caption{Comparison of feature extraction methods}
\label{table4}
\resizebox{\textwidth}{!}{%
\begin{tabular}{|c|cc|c|c|c|c|c|}
\hline
\textbf{Feature Category} & \multicolumn{2}{c|}{\textbf{Detailed  feature description}} & \textbf{Raw Features} & \textbf{PCA Features} & \textbf{Hidden Neurons} & \textbf{Train MSE} & \textbf{Validation MSE} \\ \hline
\multirow{3}{*}{\begin{tabular}[c]{@{}c@{}}Statistical\\ Features\end{tabular}} & \multicolumn{2}{c|}{Entire band} & 38 & 38 & 55 & 0.042 & 0.043 \\ \cline{2-8} 
 & \multicolumn{2}{c|}{Characteristic bands} & 114 & 100 & 40 & 0.027 & 0.028 \\ \cline{2-8} 
 & \multicolumn{2}{c|}{Feature aggregation} & 152 & 140 & 50 & 0.025 & 0.026 \\ \hline
\multirow{12}{*}{\begin{tabular}[c]{@{}c@{}}Signal  \\ representation\\ features\end{tabular}} & \multicolumn{1}{c|}{\multirow{6}{*}{\begin{tabular}[c]{@{}c@{}}Polynomial\\ basis\end{tabular}}} & 3$^{rd}$-20 & 1356 & 800 & 15 & 0.011 & \textbf{0.013} \\ \cline{3-8} 
 & \multicolumn{1}{c|}{} & 3$^{rd}$-50 & 540 & 500 & 20 & 0.011 & \textbf{0.013} \\ \cline{3-8} 
 & \multicolumn{1}{c|}{} & 2$^{nd}$-20 & 1107 & 700 & 20 & \textbf{0.010} & \textbf{0.013} \\ \cline{3-8} 
 & \multicolumn{1}{c|}{} & 2$^{nd}$-50 & 405 & 405 & 20 & 0.012 & 0.014 \\ \cline{3-8} 
 & \multicolumn{1}{c|}{} & 1$^{st}$-20 & 678 & 678 & 20 & \textbf{0.010} & \textbf{0.013} \\ \cline{3-8} 
 & \multicolumn{1}{c|}{} & 1$^{st}$ -50 & 270 & 200 & 25 & 0.015 & 0.016 \\ \cline{2-8} 
 & \multicolumn{1}{c|}{\multirow{2}{*}{\begin{tabular}[c]{@{}c@{}}Sinusoidal\\ basis\end{tabular}}} & 20 & 1356 & 1356 & 70 & 0.014 & 0.017 \\ \cline{3-8} 
 & \multicolumn{1}{c|}{} & 50 & 540 & 540 & 25 & 0.017 & 0.019 \\ \cline{2-8} 
 & \multicolumn{1}{c|}{\multirow{2}{*}{\begin{tabular}[c]{@{}c@{}}Exponential \\ basis\end{tabular}}} & 20 & 1356 & 1356 & 25 & 0.013 & 0.016 \\ \cline{3-8} 
 & \multicolumn{1}{c|}{} & 50 & 540 & 540 & 30 & 0.017 & 0.019 \\ \cline{2-8} 
 & \multicolumn{1}{c|}{\multirow{2}{*}{\begin{tabular}[c]{@{}c@{}}Power \\ basis\end{tabular}}} & 20 & 1017 & 1017 & 25 & 0.015 & 0.016 \\ \cline{3-8} 
 & \multicolumn{1}{c|}{} & 50 & 405 & 405 & 25 & 0.019 & 0.020 \\ \hline
\end{tabular}%
}
\end{table}

By examining the information in Table~\ref{table4}, we observe that in the case of statistical features, the sets obtained from the characteristic bands perform much better than the set from the whole spectrum range. This is not surprising, since the features extracted from the characteristic bands provide more details. Moreover, the aggregation of statistical features further improves performance, which confirms the hypothesis that the information contained in the two statistical feature groups is distinct and complementary. However, overall, the performance of statistical features is substantially inferior to that of representation-based features. This could be partially due to the number of extracted features; however, a more reasonable explanation is that statistical features are not as powerful as representation-based features in terms of capturing the local shape changes of the emission spectrum. This can be appreciated by observing that the best validation performance acquired by feature aggregation is merely 0.026, while the validation performance acquired by the lowest performing approximation, i.e., polynomial fitting of 1$^{st}$-50 is 0.016, which offer an improvement of approximately 38\%, with a modest increase of only 60 additional features. 

By analyzing the performance of the signal representation-based approach, we observe that polynomial basis approximation provides better results, i.e., a lower training and validation MSE than any other basis functions. For instance, the lowest performing polynomial fit model, i.e., 1$^{st}$ order with a window of 50 inputs, is comparable with the top performing sinusoidal model, 1$^{st}$ order with 20 samples, however, it utilizes about one sixth of the number of features. Indeed, the use of higher-order models and smaller window length leads to better performance, since they tend to provide a finer-grained approximation of the spectrum. However, this boosting effect tends to saturate. As shown in the Table, third-order polynomial fitting with a window size of 50 demonstrates negligible performance improvement, compared to those with a window size of 20, although the former results to a smaller number of features. The reason behind this may be that as the spectrum is divided into a larger number of smaller length windows, fitted with polynomial models, there is an increase in redundancy, which is indicated by the comparison in the number of features before and after the application of PCA. By analyzing the performance of the various approximation models, it can be concluded that third-order polynomial fitting with a window size of 50 offers a good compromise between feature complexity and performance.
\subsubsection{Machine Learning model performance}
Following the feature performance analysis described in the previous section, the group of features obtained through the application of third-order polynomial approximation with a window length of 50 samples was used to train a variety of machine learning models. To reach convergence quickly, all features were normalized to the range of (-1,1), with the outputs (temperatures of segments) normalized in the range of (0, 1). Grid search was performed on the number of features, by varying the amount of features in multiples of one hundred. The temperature values were normalized in the range of (0,1) by using the boundary values in each segment. The performance of the models is summarized in Table~\ref{table5}. In addition to the MSE on the normalized temperature values, the root of the MSE (RMSE), Mean of Relative Error (RE), Relative RMSE (RRMSE) and the Pearson correlation coefficient (R) were assessed on the original temperature values by reversing the estimated normalized temperatures to the original range.

\begin{table}[]
\centering
\caption{Comparison of machine learning models performance}
\label{table5}
\resizebox{\textwidth}{!}{%
\begin{tabular}{|c|c|c|cc|cccc|}
\hline
\multirow{2}{*}{\textbf{Machine Learning Model}} & \multirow{2}{*}{\textbf{Subclass}} & \multirow{2}{*}{\textbf{\begin{tabular}[c]{@{}c@{}}Feature number\\  \\ \end{tabular}}} & \multicolumn{2}{c|}{\textbf{Normalized Temperature}} & \multicolumn{4}{c|}{\textbf{Original Temperature (test set)}} \\ \cline{4-9} 
 & & & \multicolumn{1}{c|}{\textbf{Train MSE}} & \textbf{Test MSE} & \multicolumn{1}{c|}{\textbf{RMSE (K)}} & \multicolumn{1}{c|}{\textbf{RE}} & \multicolumn{1}{c|}{\textbf{RRMSE}} & \textbf{R} \\ \hline
MLP & - & 500 & \multicolumn{1}{c|}{0.011} & 0.0129 & \multicolumn{1}{c|}{68.3} & \multicolumn{1}{c|}{0.018} & \multicolumn{1}{c|}{0.027} & 0.993 \\ \hline
RBFN & - & 500 & \multicolumn{1}{c|}{0.014} & 0.0160 & \multicolumn{1}{c|}{75.9} & \multicolumn{1}{c|}{0.020} & \multicolumn{1}{c|}{0.030} & 0.992 \\ \hline
\multirow{3}{*}{SVR} & Linear kernel & 540 & \multicolumn{1}{c|}{0.015} & 0.0161 & \multicolumn{1}{c|}{76.2} & \multicolumn{1}{c|}{0.021} & \multicolumn{1}{c|}{0.030} & 0.992 \\ \cline{2-9} 
 & Gaussian kernel & 540 & \multicolumn{1}{c|}{0.005} & 0.0144 & \multicolumn{1}{c|}{72.2} & \multicolumn{1}{c|}{0.019} & \multicolumn{1}{c|}{0.028} & 0.993 \\ \cline{2-9} 
 & Polynomial kernel & 540 & \multicolumn{1}{c|}{0.006} & 0.0148 & \multicolumn{1}{c|}{73.1} & \multicolumn{1}{c|}{0.020} & \multicolumn{1}{c|}{0.029} & 0.992 \\ \hline
\multirow{5}{*}{GPR} & Exponential kernel & 300 & \multicolumn{1}{c|}{\textbf{2E-07}} & 0.0165 & \multicolumn{1}{c|}{77.2} & \multicolumn{1}{c|}{0.021} & \multicolumn{1}{c|}{0.030} & 0.991 \\ \cline{2-9} 
 & Rational quadratic kernel & 500 & \multicolumn{1}{c|}{\textbf{2E-07}} & 0.0159 & \multicolumn{1}{c|}{75.9} & \multicolumn{1}{c|}{0.021} & \multicolumn{1}{c|}{0.030} & 0.992 \\ \cline{2-9} 
 & Matern 32 kernel & 200 & \multicolumn{1}{c|}{4E-05} & 0.0193 & \multicolumn{1}{c|}{83.4} & \multicolumn{1}{c|}{0.023} & \multicolumn{1}{c|}{0.033} & 0.990 \\ \cline{2-9} 
 & Matern 52 kernel & 200 & \multicolumn{1}{c|}{4E-07} & 0.0193 & \multicolumn{1}{c|}{83.5} & \multicolumn{1}{c|}{0.024} & \multicolumn{1}{c|}{0.033} & 0.990 \\ \cline{2-9} 
 & Square exponential kernel & 300 & \multicolumn{1}{c|}{9E-07} & 0.0224 & \multicolumn{1}{c|}{89.9} & \multicolumn{1}{c|}{0.027} & \multicolumn{1}{c|}{0.035} & 0.988 \\ \hline
\multirow{2}{*}{Tree-based model} & LSBoost & 540 & \multicolumn{1}{c|}{0.022} & 0.0312 & \multicolumn{1}{c|}{106.1} & \multicolumn{1}{c|}{0.031} & \multicolumn{1}{c|}{0.042} & 0.984 \\ \cline{2-9} 
 & Random forest & 540 & \multicolumn{1}{c|}{0.014} & 0.0397 & \multicolumn{1}{c|}{119.6} & \multicolumn{1}{c|}{0.037} & \multicolumn{1}{c|}{0.047} & 0.979 \\ \hline
\multirow{3}{*}{Blending model} & Linear Blender &- & \multicolumn{1}{c|}{0.012} & 0.0119 & \multicolumn{1}{c|}{65.6} & \multicolumn{1}{c|}{\textbf{0.017}} & \multicolumn{1}{c|}{0.026} & \textbf{0.994} \\ \cline{2-9} 
 & Heavy Blender &- & \multicolumn{1}{c|}{0.012} & 0.0116 & \multicolumn{1}{c|}{64.7} & \multicolumn{1}{c|}{\textbf{0.017}} & \multicolumn{1}{c|}{\textbf{0.025}} & \textbf{0.994} \\ \cline{2-9} 
 & Light Blender &- & \multicolumn{1}{c|}{0.012} & \textbf{0.0115} & \multicolumn{1}{c|}{\textbf{64.3}} & \multicolumn{1}{c|}{\textbf{0.017}} & \multicolumn{1}{c|}{\textbf{0.025}} & \textbf{0.994} \\ \hline
\end{tabular}%
}
\end{table}

Training was carried out in two phases. In the first phase, the single-stage models, i.e., MLP, RBF, SVR, GPR and tree-based models were trained and their performance was assessed. The number of hidden neurons in the MLP and RBFN were optimized by grid search. The MLP network was configured to have a single hidden layer, and the hidden neuron number was optimized in the range of (5,100) units, and the optimal value was found to be 20 hidden neurons. In the case of the RBFN, a spread of 1 was selected, with at most 1500 neurons in the hidden layer. The hyperparameters of GPR, SVR, random forest and LSBoost were optimized using Bayes optimization. Early stopping was applied to prevent overfitting. Every model was trained five times to decrease variance, and the models, which performed best on the validation set were selected to be assessed on the test set. Next, attention was turned to the two-stage models, i.e., blending models, in order to further improve the ability to obtain spatially resolved temperature measurements from the spectral profiles.

Among the single-stage models, as shown in Table~\ref{table5}, the MLP performs the best. With the exception of tree-based models, almost all models have comparable performance to that of the MLP, which indicates that the selected features are effective and suiTablefor almost all the algorithms tested in this research.

In order to further improve the prediction performance, the use of model blending was explored. Three variations of the blending model were trained, namely, linear, heavy, and light blender. Their main differences are as follows: (1) Composition of weak learners: Linear blender and heavy blender use all of the single-stage models as weak learners, whereas the light blender model drops tree-based models because of their lower testing performance. (2) Selection of meta learner: Linear blender utilizes ordinary Least Squares (OLS) estimation, while both heavy and light blenders employ a MLP. 

As shown in Table~\ref{table5}, using a blending method results to performance improvements. All three blending models perform better than the MLP model with respect to the metrics of Table~\ref{table5}. However, it is also noted that this comes at the expense of increased computational and storage costs, as the blending models require the supports of a large number of weak learners. It is also shown that the MLP is a better meta learner than OLS, since the linear blender delivers a lower performance, which is not surprising, due to its linear nature, unlike the nonlinear processing capabilities of the MLP. Moreover, elimination of the lower performing tree-based models is beneficial, as it results to the light blender achieving the best performance.

\begin{figure}[hbt!]
\centering
\includegraphics[width=1.0\textwidth]{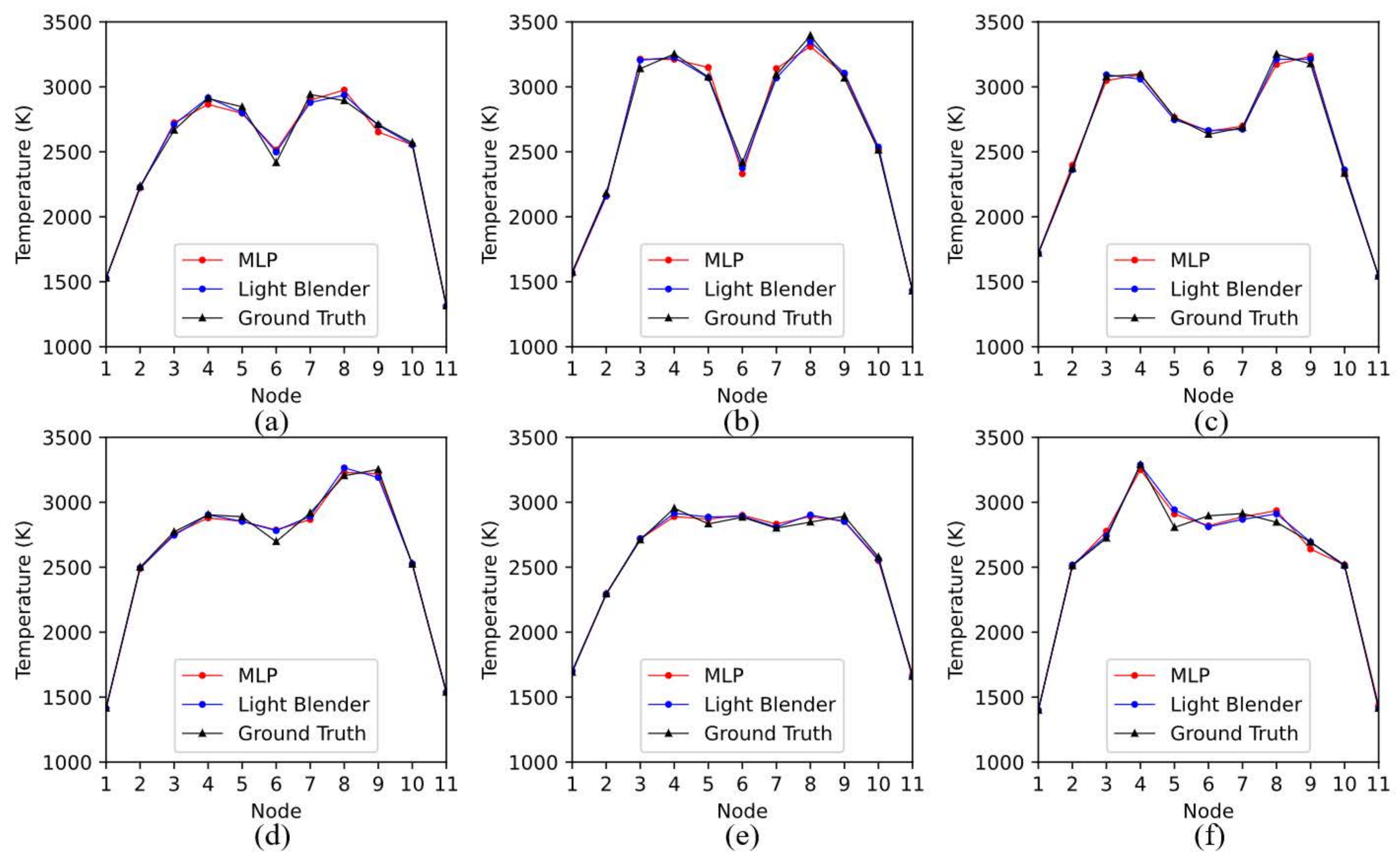}
\caption{Performance of MLP and Blender light model in recovering non-uniform temperature profiles. (a) Dual-peak Gaussian profile with a relatively small temperature variation;(b) Dual-peak Gaussian profile with a relatively large temperature variation; (c) Irregular dual-peak Gaussian profile with two “flatlands”; (d) Temperature profile with an increasing trend; (e) Trapezoidal temperature profile, (f) Temperature profile with a rush peak}
\label{fig8}
\end{figure}

Figure~\ref{fig8} displays representative temperature profile estimation examples on the test set, provided by the MLP and the Light Blender model. Both models work well, although their performance slightly degrades, when dealing with gradually irregular profiles, as shown in Figures8 (e)-(f). 

\subsection{Application of Convolutional Neural Networks on raw spectra}

Eleven CNN models were trained with the raw spectral measurements as input, which includes VGG series, Resnet series, Inceptron series, shuffle net and Squeeze net. Among them, VGG series \cite{simonyanVeryDeepConvolutional2015} are the most classical CNN architectures, which uses purely stacked convolutional layers, while ResNet \cite{heDeepResidualLearning2016} adds the skip connection to achieve the concept of residual learning, Inception \cite{cholletXceptionDeepLearning2017,szegedyGoingDeeperConvolutions2014} uses convolution kernels of varying sizes to sense diverse receptive fields, while Xception, the ultimate evolution of Inceptron, Shuffle net \cite{zhangShuffleNetExtremelyEfficient2017} and Squeeze net \cite{iandolaSqueezeNetAlexNetlevelAccuracy2016} are light level networks with more modern designs. 

In the training, Adam \cite{kingmaAdamMethodStochastic2017} was used as the optimization method, and a weight decay of 1e-4 was used to alleviate overfitting. The learning rate was preliminarily determined by the method suggested in \cite{smithCyclicalLearningRates2017}, however, it was further decreased as the selected deep learning models were prone to overfitting the training set, particularly, after removing the dropout and normalization operations, required for the regression application. According to such an operation, the learning rate was configured and varied for the various CNN architectures. In order to better navigate the training direction, the warm-up operation ~\cite{gotmareCloserLookDeep2018} was used to train the networks in the first ten epochs, during which, the learning rate was gradually increased from a tiny value to the selected learning rate. Such a soft activation of training supports the appropriate choice of training direction. During the normal training process, the Cosine Annealing learning rate schedule ~\cite{gotmareCloserLookDeep2018} was used, which helps to avoid falling into local minima.

The test performances of the CNN models are summarized in Table~\ref{table6}. Overall, all CNN models have similar performance, so that, to some extent, the results can represent the performance level, which can be reached by conventional CNN architectures with raw spectra as inputs. Strictly speaking, the best performance based on the metrics of Table~\ref{table6} is achieved by Resnet18. However, when compared to the results obtained from the combination of feature engineering and classical machine learning methods, the temperature measurement estimates using CNN are worse than most of classical machine learning models. Specifically, when comparing Tables~\ref{table5} and ~\ref{table6} , it is apparent that the values of the RMSE, RE and RRMSE metrics for Resnet18 are almost double compared to those of the light blender model.

\begin{table}[]
\centering
\caption{Performance comparison of CNN architectures}
\label{table6}
\begin{tabular}{|c|c|cc|cccc|}
\hline
\multirow{2}{*}{\textbf{CNN Model}} & \multirow{2}{*}{\textbf{Subclass}} & \multicolumn{2}{c|}{\textbf{Normalized Temperature}} & \multicolumn{4}{c|}{\textbf{Original Temperature}} \\ \cline{3-8} 
 & & \multicolumn{1}{c|}{\textbf{Train MSE}} & \textbf{Test MSE} & \multicolumn{1}{c|}{\textbf{RMSE}} & \multicolumn{1}{c|}{\textbf{RE}} & \multicolumn{1}{c|}{\textbf{RRMSE}} & \textbf{R} \\ \hline
\multirow{4}{*}{VGG} & VGG11 & \multicolumn{1}{c|}{0.032} & 0.037 & \multicolumn{1}{c|}{114.3} & \multicolumn{1}{c|}{0.033} & \multicolumn{1}{c|}{0.045} & 0.981 \\ \cline{2-8} 
 & VGG13 & \multicolumn{1}{c|}{0.032} & 0.036 & \multicolumn{1}{c|}{113.7} & \multicolumn{1}{c|}{0.033} & \multicolumn{1}{c|}{0.045} & 0.981 \\ \cline{2-8} 
 & VGG16 & \multicolumn{1}{c|}{0.033} & 0.038 & \multicolumn{1}{c|}{116.9} & \multicolumn{1}{c|}{0.034} & \multicolumn{1}{c|}{0.046} & 0.980 \\ \cline{2-8} 
 & VGG19 & \multicolumn{1}{c|}{0.040} & 0.043 & \multicolumn{1}{c|}{124.1} & \multicolumn{1}{c|}{0.037} & \multicolumn{1}{c|}{0.049} & 0.978 \\ \hline
\multirow{3}{*}{Resnet} & Resnet18 & \multicolumn{1}{c|}{\textbf{0.026}} & \textbf{0.033} & \multicolumn{1}{c|}{\textbf{108.9}} & \multicolumn{1}{c|}{\textbf{0.031}} & \multicolumn{1}{c|}{\textbf{0.043}} & \textbf{0.983} \\ \cline{2-8} 
 & Resnet34 & \multicolumn{1}{c|}{0.031} & 0.035 & \multicolumn{1}{c|}{111.0} & \multicolumn{1}{c|}{0.032} & \multicolumn{1}{c|}{0.044} & 0.982 \\ \cline{2-8} 
 & Resnet50 & \multicolumn{1}{c|}{0.028} & 0.033 & \multicolumn{1}{c|}{109.4} & \multicolumn{1}{c|}{0.032} & \multicolumn{1}{c|}{0.043} & 0.983 \\ \hline
\multirow{2}{*}{Inception} & V3 & \multicolumn{1}{c|}{0.033} & 0.036 & \multicolumn{1}{c|}{113.4} & \multicolumn{1}{c|}{0.033} & \multicolumn{1}{c|}{0.044} & 0.981 \\ \cline{2-8} 
 & Xception & \multicolumn{1}{c|}{0.032} & 0.035 & \multicolumn{1}{c|}{112.9} & \multicolumn{1}{c|}{0.033} & \multicolumn{1}{c|}{0.044} & 0.982 \\ \hline
Shuffle net & - & \multicolumn{1}{c|}{0.032} & 0.036 & \multicolumn{1}{c|}{113.4} & \multicolumn{1}{c|}{0.033} & \multicolumn{1}{c|}{0.044} & 0.981 \\ \hline
Squeeze net & - & \multicolumn{1}{c|}{0.034} & 0.039 & \multicolumn{1}{c|}{119.2} & \multicolumn{1}{c|}{0.035} & \multicolumn{1}{c|}{0.047} & 0.979 \\ \hline
\end{tabular}%
\end{table}

\begin{figure}[hbt!]
\centering
\includegraphics[width=1.0\textwidth]{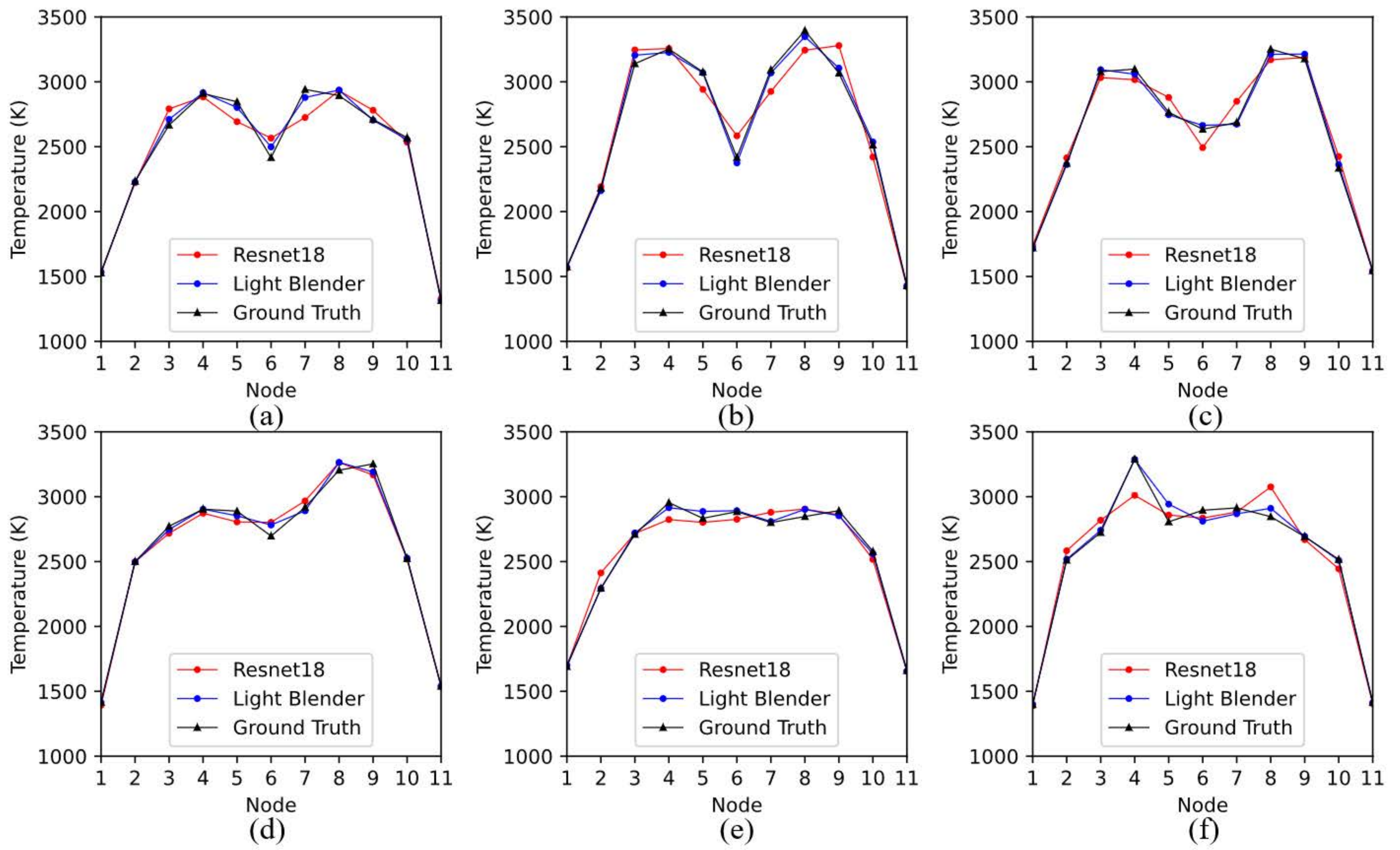}
\caption{Comparison between Resnet18 and the light blender models. (a) Dual-peak Gaussian profile with a relatively small temperature variation;(b) Dual-peak Gaussian profile with a relatively large temperature variation; (c) Irregular dual-peak Gaussian profile with two “flatlands”; (d) Temperature profile with an increasing trend; (e) Trapezoidal temperature profile, (f) Temperature profile with a rush peak.}
\label{fig9}
\end{figure}

A more direct comparison between Resnet18 and the light blender is provided in Figure~\ref{fig9}. Although Resnet18 can satisfactorily capture the basic features of the morphology of temperature profiles, it cannot provide as accurate temperature estimations as the light blender model. This comparison indicates that effective feature engineering, which is often replaced by the built-in feature extraction of CNN in the state-of-the-art ~\cite{weimerDesignDeepConvolutional2016,liuEfficientExtractionDeep2021,wiatowskiMathematicalTheoryDeep2018}, can enable classical machine learning algorithms to achieve competitive performance. 

\subsection{Comparison with state-of-the-art}
The methods described in \cite{cieszczykDeterminationPlumeTemperature2015,renMachineLearningApplied2019} were applied to the dataset used in this study. To the best of the authors’ knowledge, these are the only two studies where data-driven models are used to measure spatially resolved temperature profiles from emission spectroscopy. Both methods employ the MLP model. In \cite{cieszczykDeterminationPlumeTemperature2015}, the conventional single hidden layer architecture is employed, whereas in \cite{renMachineLearningApplied2019}, three hidden layers are used. In this work, the MLP architectures of these studies were precisely replicated. We performed grid search to fine-tune the hyperparameters of the associated MLP models. In terms of feature engineering, no feature engineering is used in \cite{renMachineLearningApplied2019}, rather, the raw spectra data is directly fed to the MLP. In \cite{cieszczykDeterminationPlumeTemperature2015}, the spectrum is divided into windows of fixed length, and the ratios of line intensities at the two sides of the spectrum are used as features. Two window lengths of 10 and 20 samples were explored in this work.

\begin{figure}[hbt!]
\centering
\includegraphics[width=1.0\textwidth]{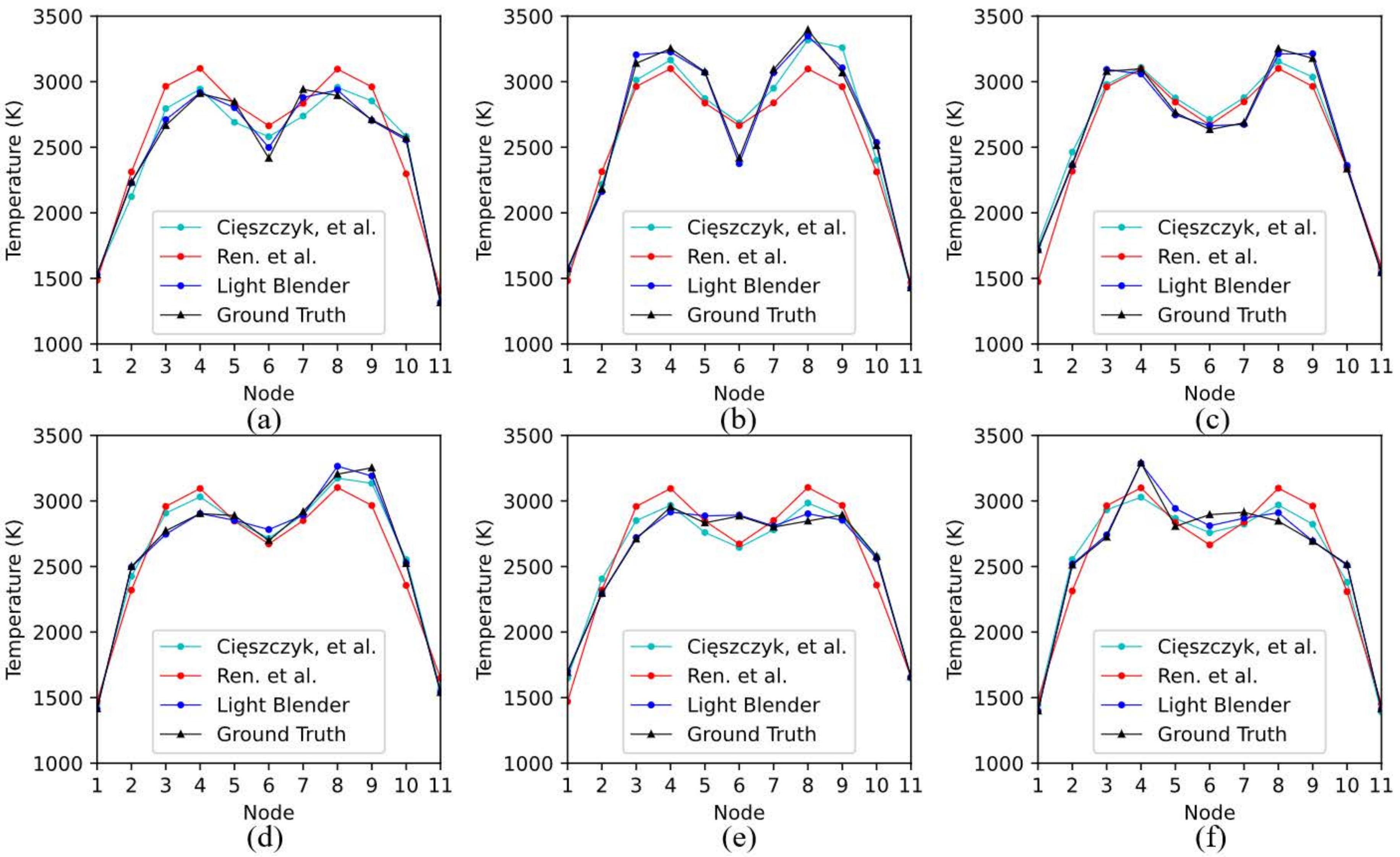}
\caption{Comparation between the state-of-the-art approaches of \cite{cieszczykDeterminationPlumeTemperature2015,renMachineLearningApplied2019} and the light blender model. (a) Dual-peak Gaussian profile with a relatively small temperature variation;(b) Dual-peak Gaussian profile with a relatively large temperature variation; (c) Irregular dual-peak Gaussian profile with two “flatlands”; (d) Temperature profile with an increasing trend; (e) Trapezoidal temperature profile, (f) Temperature profile with a rush peak.}
\label{fig10}
\end{figure}

In summary, three MLP models, two with the setup of \cite{cieszczykDeterminationPlumeTemperature2015}, and one with the architecture of \cite{renMachineLearningApplied2019}, were developed and trained. As demonstrated in Table~\ref{table7}, the light blender model (see Table~\ref{table5}) offers significant performance improvements, compared to these three models. A visual comparison of the performance of the two methods compared to the light blender model is given in Figure~\ref{fig10}, where it can be observed that, in general, the methods of \cite{cieszczykDeterminationPlumeTemperature2015} and \cite{renMachineLearningApplied2019} capture the properties of the dual-peak Gaussian profiles, which constitute the majority of the training set, but have very limited adaptability to irregular profiles, although the method in \cite{cieszczykDeterminationPlumeTemperature2015} is slightly better.

\begin{table}[]
\centering
\caption{Performance of the state-of-the-art methods}
\label{table7}
\begin{tabular}{|c|c|c|c|c|c|c|}
\hline
\textbf{Method} & \textbf{Model} & \textbf{Input} & \textbf{RMSE} & \textbf{RE} & \textbf{RRMSE} & \textbf{R} \\ \hline
\cite{renMachineLearningApplied2019} & \begin{tabular}[c]{@{}c@{}}MLP \\ (Three hidden layers)\end{tabular} & Raw spectrum & 165.7 & 0.055 & 0.065 & 0.96 \\ \hline
\multirow{2}{*}{\cite{cieszczykDeterminationPlumeTemperature2015}} & MLP & \begin{tabular}[c]{@{}c@{}}Intensity   ratios, \\ window length \\ of 50 samples)\end{tabular} & 130.5 & 0.04 & 0.051 & 0.975 \\ \cline{2-7} 
 & MLP & \begin{tabular}[c]{@{}c@{}}Intensity   ratios, \\ window length \\ of 20 samples\end{tabular} & 130 & 0.04 & 0.051 & 0.976 \\ \hline
\end{tabular}%
\end{table}

It is also notable that the CNNs (see Table~\ref{table6}), which were tested in this work, offer better performance than the two state-of-the-art methods. This supports the use of CNNs with their powerful capabilities in approximating complex mappings, compared to traditional MLP models. In turn, this highlights the importance of effective feature engineering, which when combined with machine learning models, such as the one-stage MLP, can deliver better performance than CNNs.

\section{Conclusions}

In this work, we systematically explored the feasibility of using data-driven models to recover spatially resolved temperature distributions from line-of-sight emission spectroscopy data. Two types of data-driven models were considered: (a) coupling of feature engineering and classical machine learning methods, and (b) end-to-end Convolutional Neural Networks. The main conclusions which can be drawn are as follows:
\begin{enumerate}
  \item The combination of physics-guided transformation, polynomial approximation-based features, and PCA is the most effective feature extraction methodology among the feature engineering methods attempted in this research.
  \item When using the selected group of features, the light blender model, an non-linear ensemble learning model, offers the best performance among fifteen traditional machine learning models, with RMSE, RE, RRMSE and R values of 64.3, 0.017, 0.025 and 0.994, respectively. This demonstrates the feasibility and capability of data-driven models in measuring nonuniform spatial temperature distributions from emission spectroscopy data.
  \item State-of-the-art Convolutional Neural Networks offer competitive performance compared to state-of-the-art techniques, however, they are not able to match the performance of feature engineering and machine learning models. This supports the effectiveness of feature engineering in providing substantial advantages in the context of estimating spatially resolved temperature distributions from line-of-sight spectroscopy data. 
\end{enumerate}

As for further work, it is considerable that generating more diverse temperature profile datasets, so as to further expand the application scenarios and assess the limitations of the proposed method. Moreover, we will consider further opportunities for feature engineering of emission spectroscopy data to determine more effective feature sets in order to improve the accuracy of the proposed framework.


\section*{Acknowledgements}
R.K. would like to acknowledge Dr.Hongxia Li and Prof.TJ zhang from Khalifa University for their sincere help at the beginning of constructing ideas.

\bibliographystyle{unsrt}  
\bibliography{References}

\end{document}